\documentclass{article}
\usepackage{nips13submit_e,times}
\usepackage{color}
\usepackage{graphicx,caption,subcaption,paralist}
\usepackage{amsmath,amssymb} 
\usepackage{xspace}

\usepackage{microtype}

\usepackage[unicode=true,pdfusetitle,
 bookmarks=true,bookmarksnumbered=false,bookmarksopen=false,
 breaklinks=false,pdfborder={0 0 1},backref=section,colorlinks=true]
 {hyperref}

\makeatletter
\DeclareRobustCommand\onedot{\futurelet\@let@token\@onedot}
\def\@onedot{\ifx\@let@token.\else.\null\fi\xspace}

\def\eg{\emph{e.g}\onedot} 
\def\ie{\emph{i.e}\onedot}

\def\etal{\emph{et al}\onedot}
\makeatother

\iffalse
\newcommand{\todo}[1]{{\textcolor{red}{[[TODO: #1]]}}}
\newcommand{\outline}[1]{{\textcolor{red}{[[#1]]}}}
\newcommand{\commenttext}[1]{\textcolor{blue}{[[#1]]}}
\newcommand{\commentfoot}[1]{\footnote{\textcolor{red}{#1}}}
\newcommand{\commentselfoot}[2]{{\textcolor{blue}{#1}}\comment{#2}}
\newcommand{\commentselrep}[2] {{\textcolor{blue}{#1}} {\textcolor{green}{[[\textit{#2}]]}}}
\else
\newcommand{\todo}[1]{}
\newcommand{\outline}[1]{}
\newcommand{\commenttext}[1]{}
\newcommand{\commentfoot}[1]{}
\newcommand{\commentselfoot}[2]{}
\newcommand{\commentselrep}[2]{}
\fi

\title{Identifying Synapses using Deep and Wide Multiscale Recursive
  Networks}

\author{
Gary B. Huang and Stephen Plaza \\
Janelia Farm Research Campus \\ 
Howard Hughes Medical Institute \\
19700 Helix Drive, Ashburn, VA, USA \\
\texttt{\{huangg, plazas\}@janelia.hhmi.org} 
}

\nipsfinalcopy 

\begin{document}
\maketitle

\begin{abstract}

In this work, we propose a learning framework for identifying synapses
using a deep and wide multi-scale recursive (DAWMR) network,
previously considered in image segmentation applications.  We apply
this approach on electron microscopy data from invertebrate fly brain
tissue.  By learning features directly from the data, we are able to
achieve considerable improvements over existing techniques that rely
on a small set of hand-designed features.  We show that this system
can reduce the amount of manual annotation required, in both
acquisition of training data as well as verification of inferred
detections.

\end{abstract}

\section{Introduction}

The emerging field of connectomics involves determining the
connectivity between neurons in a brain.  Recent advances in electron
microscopic (EM) imaging offer unprecedented access to the neuronal
intricacies of a brain.  However, the process of extracting consumable
information (\ie, reconstruction) from a series of highly-detailed EM
images is time-consuming, typically requiring comprehensive manual
annotation.  While efforts to automatically extract neurons using
machine image segmentation has simplified annotation, other parts of
the process, like identifying physical connections between neurons via
chemical synapses, has only recently been considered by the
machine-learning community.

Discovering the connectome of an organism is a time-intensive process
and has only been done to completion in C. elegans~\cite{white1986}.
To derive the connectome for larger organisms, significant advances
must be made in automatic image analysis.  In particular,
Plaza~\etal~\cite{plaza2014} note that with advances to image
segmentation, synapse identification is becoming a significant
bottleneck.  Synapse identification is essential to accurately
determine whether two neurons are connected.  Relatively small brains,
such as in Drosophila, contain tens of millions of synapses.  If each
synapse could be identified and annotated in approximately one minute,
this would take a single annotator a staggering 5,000 work days to
complete.

To help mitigate the bottlenecks of synapse annotations, prior work
has defined a methodology whereby pre-synaptic sites are automatically
predicted using a classifier~\cite{plaza2014a}.  Due to the inaccuracy
of the predictor, subsequent manual verification was needed.  To truly
scale synapse annotation to larger volumes, better synapse prediction
is required.

In this paper, we introduce a system for automatic synapse detection
using deep learning.  Our approach is able to achieve a large
improvement in accuracy compared to existing
methods~\cite{kreshuk2011,plaza2014a}, while also requiring much less
training data.  Therefore, the immediate benefit is a dramatic
reduction in the amount of manual effort required, both in obtaining
training data and verifying putative detections.  However, we also
note that the quality of our system makes a significant step toward
achieving human-level performance, and makes it possible to consider
fully-automated annotation without any manual verification.  We
provide extensive validation against a large image dataset in the
Drosophila~\cite{plaza2014a}.

We next give some background by discussing related literature in
automatic synapse detection and deep learning.  In
Section~\ref{sec:methods}, we describe our system for detection using
DAWMR networks, and in Section~\ref{sec:exp}, we compare the results
of our method against existing work.  We conclude with a discussion on
the potential for fully-automated detection.

\todo{pictures of synapse}

\section{Background}

In this section, we review related work in automated synapse
detection and deep learning.

\subsection{Automated Synapse Detection}

Most of the existing work on automated synapse detection has used the
ilastik toolkit~\cite{ilastik}, which allows for interactive training
of Random Forest (RF) classifiers using hand-designed features.
ilastik has been applied to detecting synapses in vertebrate
(mammalian) tissue, both with isotropic resolution~\cite{kreshuk2011}
and anisotropic resolution~\cite{kreshuk2014}.  Our work focuses on
detecting specific pre-synaptic structures in invertebrate
(Drosophila) tissue.  This problem has also been addressed using
ilastik~\cite{plaza2014a}, and we compare with this existing work in
Section~\ref{sec:exp}.

Becker~\etal~\cite{becker2013} take a slightly different approach to
detecting synapses, by estimating the location and orientation of the
synaptic cleft, and then computing (hand-designed) features at
specific locations relative to the synaptic cleft.  These features are
used as input to AdaBoost for classification.  It is uncertain how
well this method would translate to synapse detection in Drosophila,
due to the dependency on being able to consistently and reliably
determine the synaptic cleft, and the polyadic nature of many synapses
in Drosophila (having multiple post-synaptic partners per pre-synaptic
site).

\subsection{Deep Learning}

The term deep learning refers to a class of algorithms in machine
learning whose models contain multiple non-linear transformations, in
distinction to traditional ``shallow'' algorithms such as support
vector machines.  A notable example of a deep learning algorithm is
the deep belief network~\cite{Hinton2006}.  Often, such algorithms can
be trained directly on unlabeled data, to perform unsupervised feature
learning, such as the convolutional deep belief
network~\cite{Lee2009icml}.  Deep learning algorithms have recently
been applied to many tasks in computer vision and achieved
state-of-the-art results, for instance, in object
recognition~\cite{krizhevsky2012} and face
verification~\cite{sun2014}.

Deep learning techniques have also been successfully applied to
problems of segmentation and boundary prediction in EM
images~\cite{jain2008natural,ciresan2012deep}.  In this paper, we make
use of Deep and Wide Multiscale Recursive (DAWMR) networks, introduced
by Huang and Jain~\cite{huang2014} for EM boundary prediction, and
additionally applied to 3d agglomeration of neural
fragments~\cite{bogovic2014}.  DAWMR networks consist of a fast
unsupervised feature learning component, extracting features at
multiple scales and pooling features spatially for larger context,
followed by supervised classification with multilayer perceptron.
These networks were demonstrated to achieve higher accuracy than a
supervised feedforward multilayer convolutional network, as well as
requiring less time to train.

Our work differs from previous applications of deep learning to
problems in EM reconstruction, as the problem of synapse detection is
at the object-level, requiring an aggregation of evidence over
multiple voxels, rather than a binary decision applied to every voxel
or every pair of neighboring fragments.  We next discuss how we adapt
DAWMR networks to the general problem of object detection.

\section{Methodology}\label{sec:methods}

DAWMR networks were designed for the problem of image labeling, in
which a label (\eg, boundary/not-boundary) is associated with every
pixel of an image or voxel of an image volume.  In contrast, the
problem of synapse detection is one of object detection: given an
image volume, one is to identify every synapse in the volume, such as
by placing a bounding box around the synapse or identifying every
voxel belonging to the synapse.  For the data used in this paper,
described below in Section~\ref{sec:exp:data}, the synaptic structures
to be detected are all of approximately the same size, and therefore
we identify each structure with a single $x,y,z$ coordinate defining
its center, and implicitly a fixed radius defining its extent.

In order to use DAWMR networks for automated synapse detection, then,
there are two main issues that must be addressed: how to use
object-level ground-truth to train a voxel-wise classifier, and how to
produce object-level predictions from voxel-wise output.

\subsection{Synapse Voxel-wise Training}\label{sec:methods:voxel_training}

At training time, we are given an image volume and a set of
coordinates $\{s_1, \ldots, s_N\}$ indicating the center of every
synapse within the volume.  We use a simple strategy for forming
voxel-wise training data given this object-level supervision: a voxel
at spatial position $i$ has an associated label $l_i = 1$ if its
distance to any synapse at position $s_j$ is less than a predefined
threshold $r_l$, \eg $\exists j : d(i,s_j) \leq r_l$, and $l_i = 0$
otherwise.

Given the typical density of synapses within a volume, and reasonable
values for $r_l$, this training strategy leads to a large class
imbalance between positive and negative examples.  DAWMR networks are
typically trained using balanced sampling of class examples, which in
this case amounts to using all of the examples with a positive label,
and subsampling from the examples with a negative example.

\subsection{Object Prediction}\label{sec:methods:obj_prediction}

After training a DAWMR network or another voxel-wise classifier, we
can perform inference on a new image volume, producing a real-valued
prediction $p_i$ at each spatial position $i$.  For automated object
detection, we next need to transform these voxel-level predictions to
a set of coordinates indicating the set of predicted synapse centers.

To do so, we propose a straightforward method of averaging followed by
simple non-maximum suppression (NMS).  We first convolve the set of
predictions $\{p_i\}$ with an averaging filter of radius $r_a$ to
produce averaged predictions $\{a_i\}$.  We then perform an iterative
procedure where we select $j = \arg\max_i a_i$, predict a synapse at
location $j$ with confidence $a_j$, remove from consideration all $i$
within some set distance of $j$ (\ie, $a_i$ set to 0 if $d(i,j) \leq
r_n$), and then repeat.  A final set of predicted synapses can be
obtained by thresholding at a particular confidence value.

\section{Experiments}\label{sec:exp}

We compare our approach against a commonly-used baseline for automated
synapse detection.  In this section, we first give details on the data
set used for comparison, then show results in terms of
precision-recall.

\subsection{Data and Evaluation}\label{sec:exp:data}

We evaluate synapse detection algorithms on a subset of the data used
in Plaza~\etal~\cite{plaza2014a}.  This electron microscopy (EM) data
was imaged from seven medulla columns of the Drosophila optic lobe.
The image volume, preduced with FIB-SEM imaging, is of nearly
isotropic, $10^3$ nm per voxel resolution.

The specific task we consider is to identify pre-synaptic sites in the
Drosophila, often called T-bars due to a pedestal and platform
structure forming a T-like shape.  20 image volumes of size $520^3$
voxels were used for testing, containing just over 5,000 T-bars total.
A single $520^3$ voxel image volume was used for training.

We compute detection performance in terms of precision and recall.  In
order for a predicted synapse detection to be counted as a true
positive, we require that it be within a certain distance of a
ground-truth synapse location.  For our experiments, we set this
distance to be 30 voxels (300 nm), approximately equal to the width of
a T-bar structure.

\subsection{Results}

We compare results against the automatic T-bar detection algorithm
presented by Plaza~\etal~\cite{plaza2014a}, which uses the Random
Forest (RF) voxel-wise classifier from the ilastik
toolkit~\cite{ilastik}.  This work used a $K$-means based
post-processing algorithm to produce object predictions from
voxel-wise output.  We first compare this method with the averaging
and non-maximum suppression approach given in
Section~\ref{sec:methods:obj_prediction}.

The ilastik classifier is trained interactively, in an iterative
procedure where manual, voxel-level annotations are made, the
classifier trained, and the classifier output examined for whether
additional manual annotations are necessary.  As the labels are at the
voxel level, this type of supervision is much more costly in terms of
time to obtain than object-level annotations.  Therefore, we also
compare against an ilastik classifier trained statically, using the
same approach used for the DAWMR networks, presented in
Section~\ref{sec:methods:voxel_training}.

Experimental parameters were set as follows: for forming training
data, $r_l = 7$, for averaging voxel-wise predictions, $r_a = 7$, and
for NMS, $r_n = 21$.

\begin{figure}[htb]
  \centering
  \includegraphics[width=0.7\textwidth]{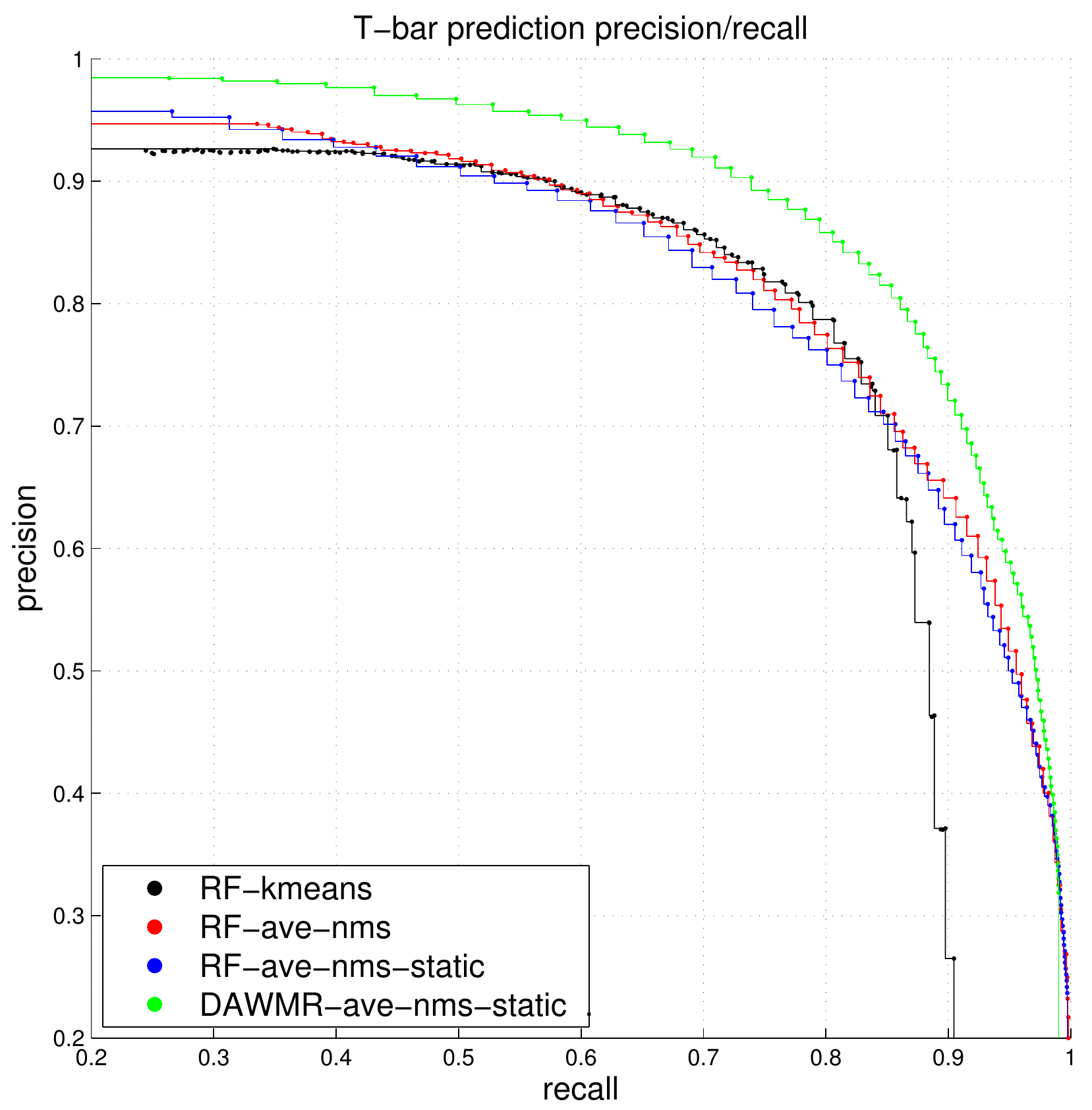}
  \caption{Precision-recall curve for automated T-bar detection.  For
    Random Forest (RF) classifiers, using averaging and non-maximum
    suppression (-ave-nms) yields higher precision at high recall than
    $K$-means for forming object predictions, and training statically
    (-static) gives comparable results to interactive training,
    despite requiring much less supervision.  The DAWMR network
    outperforms the RF methods throughout the range of recall values.}
  \label{fig:medulla_pr}
\end{figure}

Results for these methods, compared with the proposed DAWMR network
approach, are given in Figure~\ref{fig:medulla_pr}.  First, we can see
that the averaging and NMS method for producing object predictions
performs comparably to the $K$-means approach by Plaza~\etal, except
at high recall where averaging and NMS performs significantly better.
We hypothesize that this is due to the hard binarization that occurs
in order to produce connected components, prior to the $K$-means step.

Next, we see that training the Random Forest classifier statically,
using the simple transformation from object-level labels to voxel-wise
labels, gives comparable performance despite requiring much less
manual supervision and time than the typically used interactive
approach.  Lastly, the DAWMR network outperforms the Random Forest
classifiers throughout the range of recall values.  

\todo{results on MB data if time/warranted}

\section{Discussion}

Identifying synapses is a time-consuming and necessary component in
determining a connectome from EM image data.  In this work, we
introduced techniques for automated synapse detection based on deep
learning that dramatically improve precision at high recall, while
decreasing the amount of supervision required for training.

In Plaza~\etal~\cite{plaza2014a}, automated synapse detection,
thresholded to produce a recall of approximately 0.9, is used as input
to a manual verification step.  For the data set they considered, this
manual verification required approximately 350 hours.  Our proposed
DAWMR-based method doubles the precision at a 0.9 recall, therefore by
substituting this method, the time required can be reduced by half.

Unlike the problem of automatically segmenting neuronal shapes, we
believe automatic synapse detection is more tolerant of
mis-predictions, as many connections are formed from a high number of
synaptic contacts, and thus the underlying connectivity graph can stil
be recovered even if some fraction of synapses are not
detected~\cite{one_column_nature}.  Additionally, Plaza~\etal also
estimated human-to-human manual agreement to be close to 0.9 precision
at 0.9 recall.  For these reasons, our work represents a significant
step toward achieving human-level accuracy and fully-automated synapse
detection, which will considerably enhance downstream algorithms for
focusing annotation efforts and improving
segmentation~\cite{plaza2014a}.

{\small \textbf{Acknowledgements:} We thank Stuart Berg for help with
  Ilastik, the FlyEM proofreading team (Roxanne Aniceto, Lei-Ann
  Chang, Corey Fisher, Shirley Lauchie, Chris Ordish, Christopher
  Sigmund, Satoko Takemura, Julie Tran) for data annotation, and
  Toufiq Parag and Patricia Rivlin for useful discussions and
  suggestions.}

{\small{\bibliographystyle{ieee}
\bibliography{tbar_dawmr}

\begin{thebibliography}{10}\itemsep=-1pt

\bibitem{becker2013}
C.~J. Becker, K.~Ali, G.~Knott, and P.~Fua.
\newblock Learning context cues for synapse segmentation.
\newblock {\em IEEE Transactions on Medical Imaging}, 32:1864--1877, 2013.

\bibitem{bogovic2014}
J.~A. Bogovic, G.~B. Huang, and V.~Jain.
\newblock Learned versus hand-designed feature representations for 3d
  agglomeration.
\newblock In {\em International Conference on Learning Representations}, 2014.

\bibitem{ciresan2012deep}
D.~Ciresan, A.~Giusti, L.~M. Gambardella, and J.~Schmidhuber.
\newblock Deep neural networks segment neuronal membranes in electron
  microscopy images.
\newblock In {\em Advances in Neural Information Processing Systems}, pages
  2852--2860, 2012.

\bibitem{Hinton2006}
G.~E. Hinton, S.~Osindero, and Y.-W. Teh.
\newblock A fast learning algorithm for deep belief nets.
\newblock {\em Neural Computation}, 18(7):1527--1554, 2006.

\bibitem{huang2014}
G.~B. Huang and V.~Jain.
\newblock Deep and wide multiscale recursive networks for robust image
  labeling.
\newblock In {\em International Conference on Learning Representations}, 2014.

\bibitem{jain2008natural}
V.~Jain and S.~Seung.
\newblock Natural image denoising with convolutional networks.
\newblock In {\em Advances in Neural Information Processing Systems}, pages
  769--776, 2008.

\bibitem{kreshuk2014}
A.~Kreshuk, U.~Koethe, E.~Pax, D.~D. Bock, and F.~A. Hamprecht.
\newblock Automated detection of synapses in serial section transmission
  electron microscopy image stacks.
\newblock {\em P{L}o{S} {ONE}}, 9, 2014.

\bibitem{kreshuk2011}
A.~Kreshuk, C.~N. Straehle, C.~Sommer, U.~Koethe, M.~Cantoni, G.~Knott, and
  F.~A. Hamprecht.
\newblock Automated detection and segmentation of synaptic contacts in nearly
  isotropic serial electron microscopy images.
\newblock {\em P{L}o{S} {ONE}}, 6, 2011.

\bibitem{krizhevsky2012}
A.~Krizhevsky, I.~Sutskever, and G.~E. Hinton.
\newblock Image{N}et classification with deep convolutional neural networks.
\newblock In {\em Advances in Neural Information Processing Systems}, 2012.

\bibitem{Lee2009icml}
H.~Lee, R.~Grosse, R.~Ranganath, and A.~Y. Ng.
\newblock Convolutional deep belief networks for scalable unsupervised learning
  of hierarchical representations.
\newblock In {\em International Conference on Machine Learning}, 2009.

\bibitem{plaza2014}
S.~Plaza, L.~Scheffer, and D.~Chklovskii.
\newblock Toward large-scale connectome reconstructions.
\newblock {\em Current Opinion in Neurobiology}, 2014.

\bibitem{plaza2014a}
S.~M. Plaza, T.~Parag, G.~B. Huang, D.~Olbris, M.~Saunders, and P.~K. Rivlin.
\newblock Annotating synapses in large {EM} data sets.
\newblock ar{X}iv, 2014.

\bibitem{ilastik}
C.~Sommer, C.~Straehle, U.~Koethe, and F.~A. Hamprecht.
\newblock {i}lastik: Interactive learning and segmentation toolkit.
\newblock In {\em IEEE International Symposium on Biomedical Imaging}, pages
  230--233, 2011.

\bibitem{sun2014}
Y.~Sun, X.~Wang, and X.~Tang.
\newblock Deep learning face representation from predicting 10,000 classes.
\newblock In {\em Computer Vision and Pattern Recognition}, 2014.

\bibitem{one_column_nature}
S.~Takemura, A.~Bharioke, Z.~Lu, A.~Nern, S.~Vitaladevuni, P.~K. Rivlin, W.~T.
  Katz, D.~J. Olbris, S.~M. Plaza, P.~Winston, T.~Zhao, J.~A. Horne, R.~D.
  Fetter, S.~Takemura, K.~Blazek, L.-A. Chang, O.~Ogundeyi, M.~A. Saunders,
  V.~Shapiro, C.~Sigmund, G.~M. Rubin, L.~K. Scheffer, I.~A. Meinertzhagen, and
  D.~B. Chklovskii.
\newblock A visual motion detection circuit suggested by {D}rosophila
  connectomics.
\newblock {\em Nature}, 500:175--181, 2013.

\bibitem{white1986}
J.~G. White, E.~Southgate, J.~N. Thomson, and S.~Brenner.
\newblock The structure of the nervous system of the nematode {C}aenorhabditis
  elegans.
\newblock {\em Philosophical Transactions of the Royal Society of London. B,
  Biological Sciences}, 314(1165):1--340, 1986.

\end{thebibliography}
}}

\end{document}